\documentclass[11pt,a4paper]{article}
\usepackage{amssymb}
\usepackage{url}
\usepackage[hyperref]{acl2021}
\usepackage{times}
\usepackage{latexsym}
\usepackage{graphicx}
\usepackage[ruled,vlined]{algorithm2e}
\usepackage{booktabs}

\newcommand{\wikitext}[0]{\textrm{WikiText2}}
\newcommand{\wikitexthundred}[0]{\textrm{WikiText103}}
\newcommand{\PTB}[0]{\textrm{PTB}}
\newcommand{\onebw}[0]{\textrm{1BW}}
\newcommand{\openwebtext}[0]{\textrm{OpenWebText}}
\usepackage{microtype}

\aclfinalcopy 

\title{Dispatcher: A Message-Passing Approach To Language Modelling.}

\author{Alberto Cetoli \\
  QBE Europe / 30 Fenchurch St, Billingsgate, London EC3M 3BD \\
  \texttt{alberto.cetoli@uk.qbe.com} \\}

\date{}

\begin{document}
\maketitle
\begin{abstract}
This paper proposes a message-passing mechanism to address language modelling.
A new layer type is introduced that aims to substitute self-attention for unidirectional sequence generation tasks.
The system is shown to be competitive with existing methods: 
Given $N$ tokens, the computational complexity is $\mathcal{O}(N\,\mathrm{log}N)$ and the memory complexity is $\mathcal{O}(N)$ under reasonable assumptions. 
In the end, the Dispatcher layer is seen to achieve comparable perplexity to prior results while being more efficient
\footnote{Code is made available at \url{https://github.com/fractalego/dispatcher}}
.
\end{abstract}

\section{Introduction}
The introduction of self-attention \cite{vaswani_attention} has produced a considerable surge of language models \cite{Devlin2018,liu2019roberta,NEURIPS2019_dc6a7e65,lan2020albert}.
Originally, self-attention had been envisioned as a three elements algorithm (key, query, and value) to be applied onto an encoder-decoder framework for machine translation.
It soon became evident that the Transformer architecture can successfully master the most relevant NLP tasks \cite{Devlin2018} with some marginal modifications. 
One key application of self-attention has been language generation \cite{radford2019language}, where
typically the model attempts to predict the next token given a limited window of prior elements. 
A full text can thus be generated word by word. 
Self-attention - bidirectional in nature -
needs to be masked in order to avoid backward propagation of information.

Until a few years ago recurrent models \cite{sutskever2011,graves2014generating,Merity2017PointerSM,melis2018on} outperformed every other method for language modelling and generation.
This has changed with the introduction of masked self attention (MSA) models, which have achieved the state-of-the-art in language generation, culminating in some unexpected results for multi-task zero-shot learning \cite{radford2019language} as well as intriguing few-shot abilities \cite{brown2020language}.


The main argument of this work is to show that language modelling can efficiently rely on a message passing approach to perform, proposing a method that \emph{does not leverage upon self-attention}. 
Instead, the system builds a tree-like structure of forward message passing weighed by \emph{dispatching coefficients}.

In the end, the Dispatcher architecture can generate texts as well as the original Transformer model, more efficiently.
The main contributions of the paper are to introduce the novel algorithm as well as compare perplexity to the "standard" self-attention on the task of language modelling.

\section{Model}

The original Transformer architecture is composed of a number of self-attention, skip connection, and feed-forward layers.
Given $N$ tokens, the self-attention block has a computational and memory complexity of $\mathcal{O}(N^2)$ and is therefore problematic for long sequences.
Here we propose to substitute each self-attention layer with a different algorithm.

Within the Dispatcher layer, information is pushed forward onto the next tokens in a recursive fashion.
The algorithm is given a list of embeddings as \emph{input}, with the aim to create \emph{output} embeddings that contain a mixture of the tokens that precede them, without any leakage from the tokens that follow. 
The system achieves this goal by summing the tokens with themselves shifted by a power of two, iteratively. 
Each of these steps is labelled \emph{shift and sum} in Fig. \ref{fig:tree}.

\begin{algorithm}
\label{alg:dispatcher}
\SetAlgoLined
 c $\gets$ Sigmoid(Linear1(input))\;
 c $\gets$ c $\odot$ mask\;
 V $\gets$ Linear2(input)\;
 \For{$\textrm{row} = 0 \to log_2N - 1$}{
  V $\gets$ V + c[row] $\odot$ RollRight(V, $2^{row}$)\;
 }
 output $\gets$ Linear3(V)\;
 \caption{The Dispatcher Layer Algorithm}
\end{algorithm}

In the pseudo-code shown in Alg. 1 the \emph{dispatching coefficients} are written as $c \in \mathbb{R}^{N \times \mathrm{log}_2N}$, 
whereas $V \in \mathbb{R}^{N \times d}$ is the tensor containing the hidden states used as a working memory in the main loop, with embedding dimension $d$.
The $\mathrm{Linear}$ functions are dense layers, while $\mathrm{RollRight}$ shifts the tokens to the right.

The message coming from the prior tokens follows a binary tree structure, as depicted in Fig. \ref{fig:tree}.
The sum is weighed by the \emph{dispatching coefficients}, which effectively decide whether information coming from the left of the tree should propagate further, and by what amount. These weights are computed through a dense layer applied to the original tokens.
A constant mask is applied to the tensor $c$ after it has been computed to avoid leakage after the $\mathrm{RollRight}$ operation. 

The algorithm presented above describes a single-head unit.
As with self-attention, this layer can be split into a set of Dispatcher heads to improve performance.
The number of heads then becomes another hyper-parameter to tune during training.

Finally, if the number of input embeddings is not a power of two, the loop stops when the shift value is greater than the input length.

\subsection{Dispatcher Dropout}
\label{sec:dropout}
A quick modification of Alg. 1 can introduce an effective dropout by randomly skipping a \emph{shift and sum} step in training with a probability given by a dropout value between $0$ and $1$.
Notice that in this procedure dropout makes the algorithm quicker, albeit with the same computational complexity.


\subsection{Computational complexity}
\begin{figure}[t!]
    \centering
         \includegraphics[width=0.55\textwidth]{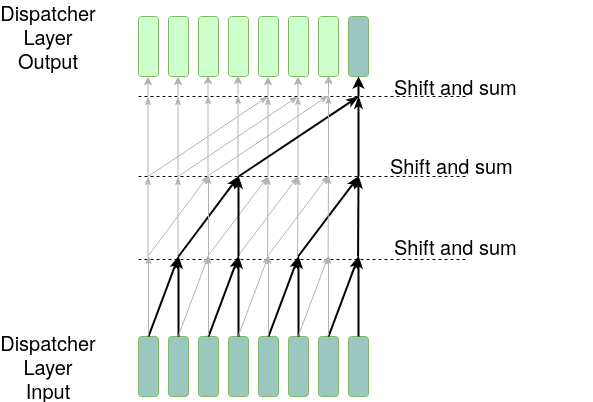}%
\caption{\label{fig:tree}
A representation of how information is passed from the input tokens to the output within the Dispatcher Layer. 
At every vertex of this directed graph the embeddings are summed together, each sum weighed by the \emph{dispatching coefficients}.
These weights determine how much of the message from the left needs to be passed onto the right.
For clarity the paths to the last output item are painted in a darker color.
}
\end{figure}

The creation of the \emph{dispatching coefficients} is linear in time, as a dense layer is a applied to every input token.
The main algorithm repeats $\mathrm{log}_2N$ times a weighted sum. If $d$ is the dimension of the embeddings, the computational complexity is $\mathcal{O}(d\,N\,\mathrm{log}N)$.

\subsection{Memory complexity}
The system computes the dispatching coefficients in every layer with a space complexity of $\mathcal{O}(N\,\mathrm{log}N)$.
In addition, the algorithm uses at every step a set of embeddings $V$ with complexity $\mathcal{O}(N \times d)$. In a typical scenario $d \gg \mathrm{log}_2N$, yielding an effective asymptotic linear memory consumption $\mathcal{O}(N \times d)$.


\section{Evaluation}
\subsection{Datasets}
The algorithm is evaluated on the following datasets: \PTB{} \cite{MikolovZweig2010}, \wikitext{} and \wikitexthundred{} \cite{Merity2017PointerSM}, and One Billion Word \cite{ChelbaMSGBK13}. A simple pre-processing step uses the special token \textrm{$<$EOS$>$} to indicate the end of each sentence.

Among the sets, \PTB{} and \wikitext{} are the smallest, with only 4.9MB and 11MB of text data for training respectively. This is to be compared to the 515MB training set of \wikitexthundred{} and 3.9GB of \onebw{}.
While the larger dataset, \onebw{} only models short-term dependency because the sentences have been shuffled.

An additional corpus called \openwebtext{} \cite{Gokaslan2019OpenWeb} is used to train the larger Dispatcher model. This set was created as an open alternative to the one used when training GPT-2 \cite{radford2019language} and consists of about 40 GB of text data.

\subsection{Training}
At first, the Dispatcher algorithm is compared against a masked self-attention model (see Sec. 4.1 and 4.2).
Rather than trying to optimize for the MSA and the Dispatcher separately, we choose an identical set of parameters (embedding size, number of heads, layers) and compare their perplexity.
While this approach might not give the best results for each model, it helps to show that the two algorithms performs similarly under similar conditions.

Secondly, a slightly bigger Dispatcher model (Sec. 4.3) is trained for \emph{one epoch only} on the \openwebtext{} corpus.
The goal is not to achieve state-of-the-art results, rather to prove that the proposed architecture can reach comparable perplexity to Transformer-based models. 
In the spirit of simplicity, we use a single head for all the models, which are trained on the same single-GPU machine.
All the models are implemented in \emph{PyTorch} \cite{NEURIPS2019_9015}.

\subsection{Masked self-attention (MSA) and Plain Dispatcher}
\begin{figure}[t!]
    \centering
         \includegraphics[width=0.5\textwidth]{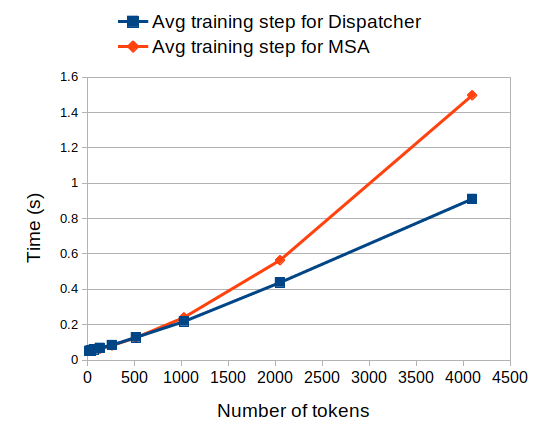}%
\caption{\label{fig:timing}
Average time in seconds for a single training step on \wikitext{} as a function of the number of input tokens.
Both models are trained on the same single-GPU instance.
The asymptotic behavior appears remarkably different. 
}         
\end{figure}

These models have embedding and inner dimension size 512, 6 layers, and only 1 head. 
The training batch size is 20 and dropout is set to 0.2, using 512 tokens.
The only difference between the two models is the self-attention/Dispatcher layer.

Tokenization is done using a pre-trained WordPiece tokenizer made available by HuggingFace \cite{wolf2020huggingfaces}.
On training and evaluation the vocabulary is further restricted on each dataset to improve performance, as a consequence the number of parameters changes depending on the relevant dataset's vocabulary size.

\subsection{Dispatcher after \openwebtext{}}
This model has a single head, 480 embedding and inner dimension size, and 12 layers with a mini-batch size of 5 and no dropout.
A BPE tokenizer - pre-trained on \openwebtext{} - is used.
The number of tokens used for this model is 1024.
First pre-trained on \openwebtext{}, the model is then fine-tuned onto the relevant sets.

\subsection{Discussion}

\begin{table*}[t!]
\centering
\begin{tabular}{c c c c c}
    \toprule
        \textbf{Model Type} & \textbf{\PTB{}} & \textbf{\wikitext{}} & \textbf{\wikitexthundred{}} & \textbf{\onebw{}}\\
    \midrule
    \textrm{Masked Self-Attention (18M / 30M / 39M / 41M)} & 40.58 & 55.05 & 22.56 & 66.52\\
    \textrm{Plain Dispatcher (17M / 27M / 36M / 38M)} & 35.40 & 50.23 & 24.39 & 53.32\\
    \midrule
    \textrm{Dispatcher after 
    \openwebtext{} (59M)} & 18.95 & 22.74 & 20.38 & 36.76 \\
    \textrm{\cite{fan2020accessing} (44M)} & - & - & 22.4 & - \\
    \textrm{\cite{lmtransformer2019} (395M)} & 31.34 & 34.11 & 20.42 & -\\
    \textrm{\cite{dai-etal-2019-transformer} (257M / 0.8B)} & - & - & 18.3 & 21.8\\
    \textrm{\cite{radford2019language} (1.5B)} & 35.7 & 18.34 & 17.48 & 42.16\\
    \textrm{\cite{shoeybi2020megatronlm} (355M)} & - & - & 19.31 & -\\
    \textrm{\cite{shoeybi2020megatronlm} (8.3B)} & - & - & 10.81 & -\\
    \bottomrule
\end{tabular}
\caption{
Top: The Dispatcher architecture is evaluated concurrently with a masked self-attention model yielding similar results. Bottom: The Dispatcher pre-trained on \openwebtext{} is compared to some recent results achieved using a variant of the Transformer architecture. 
All the results refer to the test perplexity.
\label{table:results}
}
\end{table*}
The MSA model and the Plain Dispatcher are evaluated against four different datasets, as shown in the first two rows of Table 1.
The results are quite similar, with the Dispatcher architecture seen performing better on the smaller sets \PTB{} and \wikitext{}. 
This is arguably due to the model having fewer parameters and being less prone to overfitting.
Conversely, the larger MSA model wins on \wikitexthundred{}.
The Dispatcher overtaking MSA on \onebw{} is more challenging to explain in terms of model size and seems to suggest its enhanced ability to model short-term dependencies, at least in this one-headed configuration.

A striking difference between the MSA  and the Dispatcher is however shown in Fig. \ref{fig:timing}, which plots the average time for a single training step as a function of the number of input tokens.
While the recorded times are configuration-specific, 
the asymptotic behavior looks radically different, suggesting the Dispatcher architecture as a better candidate for longer sequences.

A single epoch of training onto the \openwebtext{} dataset boosts the Dispatcher performance into competitive results for a model of this size, after fine-tuning on the relevant corpus.
This is shown in the third row of Table 1, presenting our top results. 

The rest of Table 1 is a showcase of the most recent self-attention based models. 
Notably, our results on \PTB{} and \wikitext{} are among the best in the literature, surpassing the results in \cite{lmtransformer2019} which are obtained by fine-tuning a pre-trained BERT model.
This is most likely due to the \openwebtext{} corpus being a better set for language generation than BookCorpus (used by BERT), but it bodes well for the algorithm presented here that it can compete against models with one order of magnitude more parameters. 
The last three rows relate to zero-shot results.
As one can expect, the 59M parameters Dispatcher cannot compete with models two orders of magnitude larger. 

\section{Related works}
The way information is funneled to higher layers in Fig. \ref{fig:tree} is reminiscent of convolutional neural networks (CNN) \cite{ConvTransformer,ConvSeq2SeqLearning}. It is especially evocative of \emph{dilated convolutions} as presented in \cite{pmlr-v80-oord18a}. While similar, the method presented here is not technically a convolution, which by definition requires the same operator being translated over the input elements. In this paper the dispatching coefficients are \emph{local} to the tokens.

Another way to visualize the Dispatcher algorithm is as a set of overlapping Recursive NN acting on binary trees \cite{RecursiveNNGoller,RecursiveNNSocher} which share parameters where the trees overlap.
It is however important to keep in mind that the \emph{shift and sum} iteration only performs a weighted sum of the input embeddings, achieving competing performance only when repeated within a multi-layer structure.

The computational cost of large models has become a source of concern in terms of scalability as well as energy consumption \cite{strubell-etal-2019-energy}.
For this reason, a growing number of approximations \cite{wang2020linformer,kitaev2020reformer,zaheer2021big,choromanski2020rethinking,zhai2021attention} has appeared in the literature, suggesting modifications to the main self-attention layer. 
These approximations tend to leverage linear algebra properties to speed up calculations, 
capturing the essence of the Transformer architecture into more efficient algorithms.
In many cases the approximation makes the model irreducibly bidirectional, thus hindering language generation tasks.

Finally, the concept of message passing is understood to describe Graph Convolutional Networks \cite{KipfW16,Geerts2020} and by extension the self-attention mechanism in the Transformer architecture.
The Dispatcher algorithm makes message passing explicit by keeping the routing topology constant while relying on the coefficients to distribute the message within a set of binary trees.

\section{Conclusions}
A novel architecture dedicated to language modelling is introduced and shown to achieve comparable perplexity with self-attention based models, requiring less computational and memory resources.
Pre-training onto a large dataset (\openwebtext{}) allows the model to achieve competitive performances on \PTB{}, \wikitext{}, \wikitexthundred{}, and \onebw{}. The search for faster architectures will allow a bigger context window for language modelling, while at the same time reducing the associated environmental cost.
The proposed Dispatcher layer appears as a viable candidate for improved efficiency.

Finally, low perplexity in the task of language modelling is often predictive of high-quality text generation.
This intriguing possibility will be pursued in a future work.

\bibliographystyle{acl_natbib}
\bibliography{anthology,acl2020,biblio}
\end{document}